\documentclass[10pt,twocolumn,letterpaper]{article}

\usepackage{iccv}
\usepackage{times}
\usepackage{epsfig}
\usepackage{graphicx}
\usepackage{amsmath}
\usepackage{amssymb}
\usepackage{booktabs}

\usepackage[ruled,algonl,lined]{algorithm2e}

\DeclareMathOperator*{\argmin}{argmin}
\usepackage{makecell}
\usepackage{caption}
\usepackage[pagebackref=true,breaklinks=true,letterpaper=true,colorlinks,bookmarks=false]{hyperref}

\iccvfinalcopy % *** Uncomment this line for the final submission

\def\iccvPaperID{4309} % *** Enter the CVPR Paper ID here

\ificcvfinal\fi
\begin{document}

%%%%%%%%% TITLE
\title{Deep Meta Learning for Real-Time Target-Aware Visual Tracking}

\author{Janghoon Choi\\
	ASRI, Department of ECE, \\Seoul National University\\
	{\tt\small ultio791@snu.ac.kr}
	\and
	Junseok Kwon\\
	School of CSE, \\Chung-Ang Univeristy\\
	{\tt\small jskwon@cau.ac.kr}
	\and
	Kyoung Mu Lee\\
	ASRI, Department of ECE, \\Seoul National University\\
	{\tt\small kyoungmu@snu.ac.kr}
}

\maketitle
\ificcvfinal\thispagestyle{empty}\fi

%%%%%%%%% ABSTRACT
\begin{abstract}
	In this paper, we propose a novel on-line visual tracking framework based on the Siamese matching network and meta-learner network, which run at real-time speeds. Conventional deep convolutional feature-based discriminative visual tracking algorithms require continuous re-training of classifiers or correlation filters, which involve solving complex optimization tasks to adapt to the new appearance of a target object. To alleviate this complex process, our proposed algorithm incorporates and utilizes a meta-learner network to provide the matching network with new appearance information of the target objects by adding target-aware feature space. The parameters for the target-specific feature space are provided instantly from a single forward-pass of the meta-learner network. By eliminating the necessity of continuously solving complex optimization tasks in the course of tracking, experimental results demonstrate that our algorithm performs at a real-time speed while maintaining competitive performance among other state-of-the-art tracking algorithms.
\end{abstract}

%%%%%%%%% BODY TEXT

\section{Introduction}

Visual object tracking is one of the fundamental and practical problems among the fields of computer vision research, and it has seen applications in automated surveillance, image stabilization, robotics and more. Given the initial bounding box annotation of an object, visual tracking algorithms aim to track the specified object throughout the subsequent part of the video without losing the object under various circumstances such as illumination change, blur, deformation, fast motion, and occlusion.

Recently, with the increasing use of deep learning and convolutional neural networks (CNN) \cite{lenet} in computer vision applications for their rich representation power and generalization capabilities  \cite{alexnet,frcnn,vgg}, there have been numerous studies on utilizing the rich and general feature representation of the CNNs for visual tracking task \cite{nipstrack,convcorr,mdnet,fctrack,siamis}. Most algorithms incorporate the deep convolutional features used in object recognition systems \cite{alexnet,vgg,frcnn}. On top of these feature representations, additional classifiers or correlation filters are trained for on-line adaptation to the target object \cite{mdnet,cnnsvm,fctrack,convcorr,ccot,csrdcf,deepsrdcf,MCPF}.

\begin{figure}[t]
	\begin{center}
		\includegraphics[width=1.0\linewidth]{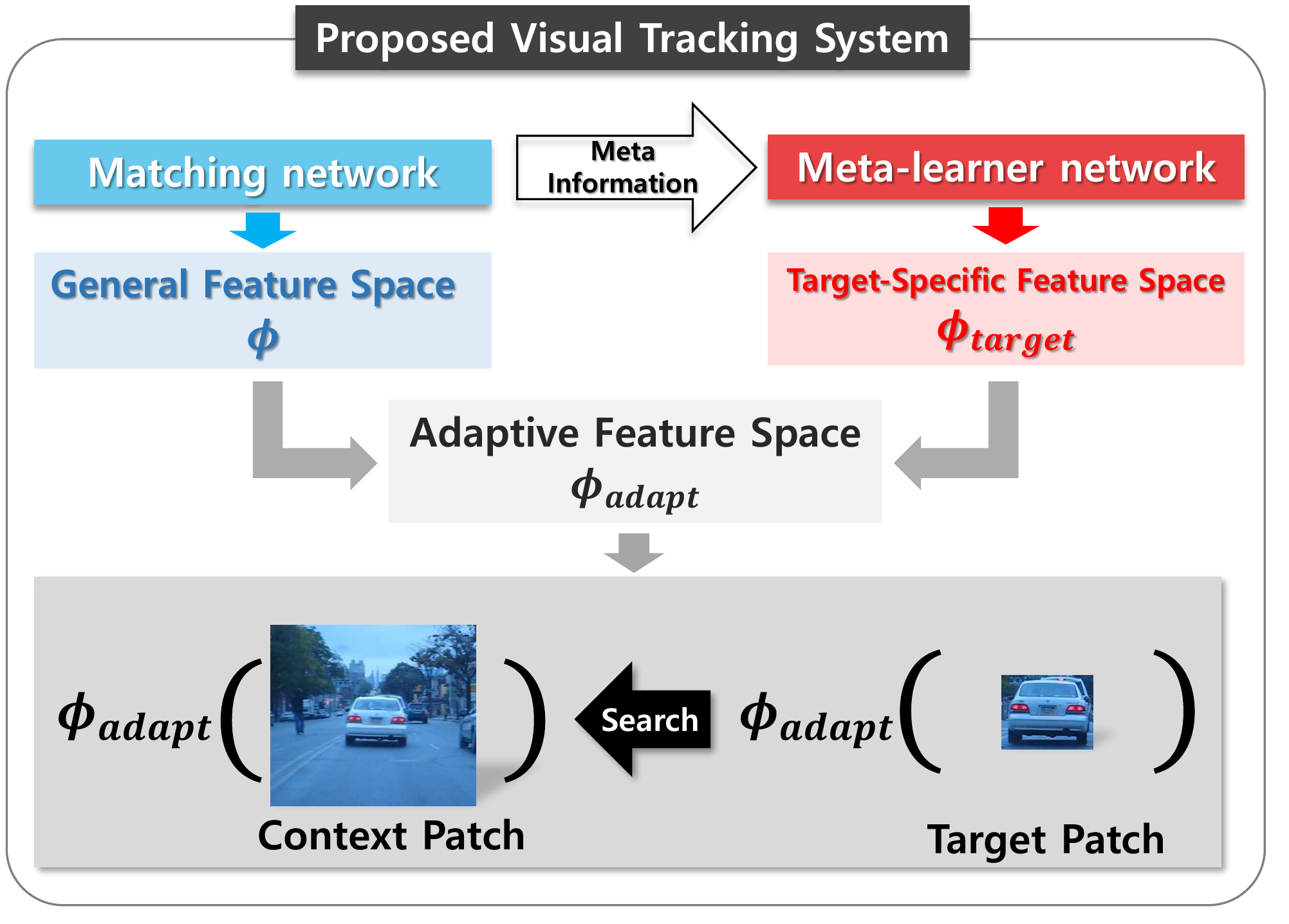}
	\end{center}
	\vspace{-5mm}
	\caption{{\textbf{Motivation of the proposed visual tracker.}} Our framework incorporates a meta-learner network along with a matching network. The meta-learner network receives meta information from the matching network and provides the matching network with the adaptive target-specific feature space needed for robust matching and tracking.}
	\label{fig:intro}
	\vspace{-5mm}
\end{figure}

While these methods were successful in obtaining high performance metrics in well-known benchmarks and datasets \cite{vot,otb} using deep representations, the majority of these algorithms were not designed as an integrated structure, where two different systems (i.e. deep feature network system and target classifier systems) are built and trained separately, not being closely associated. This causes several problems when the framework is naively applied to the problem of visual tracking, where the classifier system is in constant need of being updated in order to adapt to the appearance changes of the target object, while the number of positive samples are highly limited.

Since an update operation requires solving complex optimization problems for a given objective function using methods such as stochastic gradient descent (SGD) \cite{mdnet}, Lagrange multipliers \cite{cnnsvm}, and ridge regression \cite{cnnsvm,ccot,deepsrdcf}, most tracking algorithms with deep representations run at low speeds under $20$ fps, thus making real-time applications unrealizable. Moreover, since the updates are often achieved by utilizing a handful of target appearance templates obtained in the course of tracking, while this strategy is inevitable, classifiers are prone to overfitting and losing generalization capabilities due to insufficient positive training samples. To deal with this prevalent overfitting problem, most algorithms incorporate a hand-crafted regularization term with a training hyper-parameter tuning scheme to achieve better results.

Our approach tackles the aforementioned problems by building a visual tracking system incorporating a Siamese {\textit{matching network}} for target search and a {\textit{meta-learner network}} for adaptive feature space update. We use a fully-convolutional Siamese network structure analogous to \cite{siamfc} for searching the target object in a given frame, where target search can be done fast and efficiently using the cross-correlation operations between feature maps. For the meta-learner network, we propose a parameter prediction network inspired by recent advances in the meta learning methodology for few-shot learning problems \cite{metanet,matchos,maml}.

The proposed meta-learner network is trained to provide the matching network with additional convolutional kernels and channel attention information so that the feature space of the matching network can be modified adaptively to adopt new appearance templates obtained in the course of tracking without overfitting. The meta-learner network only sees the gradients from the last layer of the matching network, given new training samples for the appearance. We also employ a novel training scheme for the meta-learner network to maintain the generalization capability of the feature space by preventing the meta-learner network from generating new parameters that cause overfitting of the matching network. By incorporating our meta-learner network, the target-specific feature space can be constructed instantly with a single forward pass without any iterative computation for optimization and is free from innate overfitting, improving the performance of the tracking algorithm. Fig.\ref{fig:intro} illustrates the motivation of the proposed visual tracking algorithm. We show the effectiveness of our method by showing consistent performance gains in 5 different visual tracking datasets \cite{otb,LaSOT,tc128,uav123,vot} while maintaining a real-time speed of 48 fps.

\begin{figure*}
	\begin{center}
		\includegraphics[width=0.98\linewidth]{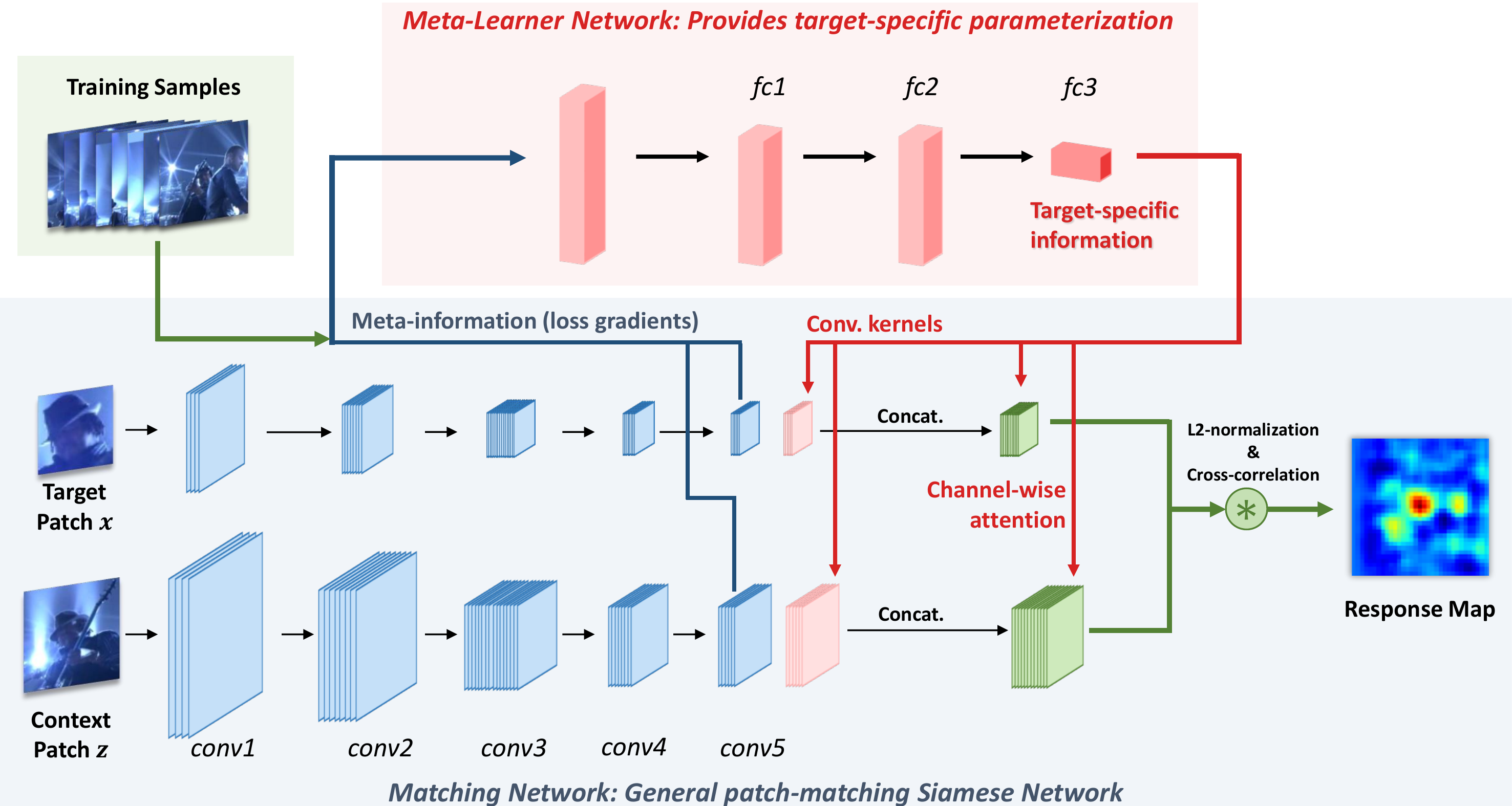}
	\end{center}
	\vspace{-2mm}
	\caption{\textbf{Overview of proposed visual tracking framework.} The matching network provides the meta-learner network with meta-information in the form of loss gradients obtained using the training samples. Then the meta-learner network provides the matching network with target-specific information in the form of convolutional kernels and channel-wise attention.}
	\label{fig:overview}
	\vspace{-5mm}
\end{figure*}

%-------------------------------------------------------------------------
\section{Related Work}
%-------------------------------------------------------------------------
\noindent \textbf{General visual tracking approaches: }
Conventional tracking algorithms can be largely grouped into two approaches. One approach builds a generative appearance model of the target based on previously observed examples. This generative model can be used to find the target in the upcoming frames by finding the region that can be best described by the model, where sparse representation and linear subspace representation is often utilized \cite{IVT,CST,MTT,L1}. The other approach aims to build a discriminative classifier to distinguish the target region from the background region. This discriminative classifier can be used to find the target region in the upcoming frames by solving a binary classification problem \cite{KCF,Muster,srdcf,meem,Struck,TLD}. Recently, correlation filters have gained great popularity among the visual tracking methods since the seminal works of \cite{mosse} and \cite{KCF} due to their simplicity and computational efficiency in the Fourier frequency domain. Many new approaches have been proposed based on the correlation filter learning framework such as color attribute features \cite{color}, using multi-resolution feature maps \cite{hdt,ccot}, accurate scale estimation \cite{dsst}, spatial regularization \cite{dsst}, and factorized convolution operators \cite{ECO}. 

%-------------------------------------------------------------------------
\noindent\textbf{Visual tracking methods using deep representations: }
With the growing popularity of the application of deep convolutional networks to a wide range of computer vision tasks, many novel visual tracking algorithms make use of the powerful representation capabilities of the convolutional neural networks (CNN). Starting with \cite{nipstrack}, where the encoder representation of the denoising autoencoder was used, \cite{mdnet} used feature representations of the VGG-M network \cite{vgg} and  \cite{fctrack} also used VGG feature maps. Many correlation filter-based tracking algorithms also utilize the powerful representation capacity of the CNN by training the correlation filters on the feature maps of the network. Recent approaches include hierarchical correlation filters \cite{convcorr}, adaptive hedging of correlation filters \cite{hdt}, continuous convolutional operators \cite{ccot}, sequential training of features \cite{stct}, and spatial regularization \cite{deepsrdcf}. Other than correlation filter-based algorithms, approaches to design an end-to-end framework for visual tracking have recently emerged. They employ the two-flow Siamese architecture networks commonly used in stereo matching \cite{mccnn} and patch-matching \cite{matchnet} problems. \cite{siamis} and \cite{100fps} trained a Siamese network to learn the two-patch similarity function that shares the convolutional representation. \cite{siamfc} proposed a more end-to-end approach to visual tracking where the Siamese network can localize an exemplar patch inside a search patch. They use a  fully convolutional architecture that adopts a cross-correlation layer which significantly lowers the computational complexity. Based on the framework of \cite{siamfc}, recent approaches incorporate triplet loss \cite{siamtriplet}, region proposal networks \cite{siamrpn}, distractor-aware features for suppressing semantic distractors \cite{dasiamrpn} and two-fold Siamese networks for semantic and appearance features \cite{sasiam}. 

%-------------------------------------------------------------------------
\noindent\textbf{Meta learning methods for few-shot image recognition task and visual tracking: }
There are recent approaches for learning to classify from a few given examples using meta learning methodologies \cite{matchos,maml,metalstm,metanet}. In \cite{matchos}, authors proposed a network architecture that employs characteristics of non-parametric nearest-neighbor models to solve \textit{N}-way, \textit{k}-shot learning tasks, where a small support set is given. \cite{maml} made use of a pre-trained network as a good initialization and then trained the meta-learner to effectively fine-tune the network based on few given examples. In \cite{metanet}, a two-level structure of meta-learner and base-learner both equipped with fast and slow weights was used. The meta-learner acquires the meta information from the base-learner in the form of loss gradients, and then provides the based-learner with fast parameterization while preserving generalization capabilities. Recently, \cite{metatracker} proposed a meta-learner based optimizer analogous to \cite{maml} for the visual tracking task, where they chose \cite{crest} and \cite{mdnet} as the baseline algorithms to show the effectiveness of their update step, decreasing the number of training iterations thus improving the speed of the baseline methods. Whereas the aim of our method is to update the network using the meta-learner with a single iteration at real-time speeds, providing the network with new adaptive kernels and feature space representation without overfitting, which can be achieved by our regularizing training scheme.

%-------------------------------------------------------------------------
\section{Tracking with Meta-Learner}

In the following subsections, we first provide an overview of our proposed visual tracking framework, where a brief explanation of our framework and the visual tracking procedure are given. Then we describe the implementation and training details for the components of our framework. 

%-------------------------------------------------------------------------
\subsection{Overview of Proposed Method} \label{sub:overview}

\subsubsection{Components}
Our framework is largely composed of two components, a matching network and a meta-learner network. The matching network is a fully-convolutional Siamese network that takes two images as inputs where $x$ is denoted as an image patch of the target and $z$ is an image patch of the larger context area that contains the target. The matching network takes these inputs, extracts the feature maps using the $N$-layer feature extractor CNN network $\phi_{{\bf{w}}}(\cdot)$, and produces the final response map $f_{\bf{w}}(x,z)$ by cross-correlation operation between the feature maps. This process can be expressed as follows,
\begin{equation}\label{eq:match}
f_{\bf{w}}(x,z)=\phi_{{\bf{w}}}(x) \ast \phi_{{\bf{w}}}(z),
\end{equation}
where $\ast$ represents the cross-correlation operator between two feature maps and ${\bf{w}}=\{w_1,w_2,...,w_N\}$ represents the set of trained kernel weights for each layer of the feature extractor CNN. To train the feature extractor CNN, we minimize a differentiable loss function given as $\ell(f_{{\bf{w}}}(x,z),y)$, where the loss function measures the inaccuracy in predictions of $f_{\bf{w}}$, given $y$ as the ground-truth response map. 

The meta-learner network provides the matching network with target-specific weights given an image patch of the target $x$ with context patches ${\bf{z}_\delta}=\{z_1,...,z_M\}$ previously obtained and cropped around the target's vicinity. To adapt the weights to the target patch, we use the averaged negative gradient $\delta$ of the loss function for the last layer of the matching network taken as, 
\begin{equation}\label{eq:grad}
\delta = \sum_{i=1}^{M} - \frac{1}{M}\frac{\partial\ell(f_{\bf{w}}(x,z_i),\tilde{y}_i)}{\partial w_N},
\end{equation}
where $\tilde{y}_i$ is the generated binary response map \textit{assuming} the target is located at the correct position inside the context patch $z_i$. 
The meta-learner network is designed based on the fact that the characteristic of $\delta$ is empirically different according to a target. 
Then, given $\delta$ as an input, the meta-learner network $g_\theta(\cdot)$ can generate target-specific weights $w^{target}$ corresponding to the input as, 
\begin{equation}\label{eq:meta}
w^{target} = g_\theta(\delta),
\end{equation}
where $\theta$ is the parameter for the meta-learner network. The new weights are used to update the matching network's original weights as in,
\begin{equation}\label{eq:update}
f_{{\bf{w}}^{adapt}}(x,z)=\phi_{{\bf{w}}^{adapt}}(x) \ast \phi_{{\bf{w}}^{adapt}}(z),
\end{equation}
where ${\bf{w}}^{adapt}=\{w_1,w_2,...,[w_N,w^{target}]\}$, concatenating $w^{target}$ to $w_N$ of the last layer for feature extraction. The meta-learner network also generates channel-wise sigmoid attention weights for each channel of the feature map to further adjust the feature representation space where the weights can be applied by channel-wise multiplication.
Fig.\ref{fig:overview} shows an overview of the proposed method.

\begin{algorithm}[t]
\small{
	\SetKwData{Left}{left}\SetKwData{This}{this}\SetKwData{Up}{up}
	\SetKwFunction{Union}{Union}\SetKwFunction{FindCompress}{FindCompress}
	\SetKwInOut{Input}{input}\SetKwInOut{Output}{output}
	
	\Input{Tracking sequence of length $L$ \\ Initial target state $\mathbf{s}_1$\\ Corresponding initial target template $x$}
	\Output{Tracked target states $\mathbf{s}_t$} \BlankLine
	
	\emph{// For every frame in a tracking sequence}\\
	\For{$t=2$ to $L$} {
		Obtain a candidate context image $z'$ based on the previous target state $\mathbf{s}_{t-1}$;\\
		Obtain a response map $\hat{y}$ using the matching network as in eq.\eqref{eq:match} or eq.\eqref{eq:update};\\
		Apply cosine window $h$ to $\hat{y}$, find the position and scale with maximum response, and obtain a new state $\mathbf{s}_t$;\\
		\underline{}\\
		\emph{// Store context image in the memory if confident}\\
		\If{$\hat{y}[\mathbf{s}_t] > \tau$} {
			Obtain new context image $z_t$ based on $\mathbf{s}_t$ and store it in the memory ${\bf{z}}_{mem}$;\\
		}
		\underline{}\\
		\emph{// Update weights every $T$ frames}\\
		\If{$(t\mod T) == 0$} {
			Choose $M$ samples ${\bf{z}_\delta}$ from memory ${\bf{z}}_{mem}$ under minimum entropy metric \eqref{eq:entropy};\\
			Obtain a loss gradient $\delta$ as in eq.\eqref{eq:grad};\\
			Obtain target-specific adaptive weights $w^{target}$ as in eq.\eqref{eq:meta}\\
			Update ${\bf{w}}^{adapt}$ for the matching network in \eqref{eq:update}\\
		}
	}
	\caption{\small{Visual tracking with meta-learner network}} \label{alg:track}
}
\end{algorithm}

%-------------------------------------------------------------------------
\subsubsection{Tracking algorithm}
Tracking is performed in a straightforward and simplistic manner to ensure fast performance. Given a target patch $x$ and its previous state, a context image $z$ in a new frame can be cropped based on the previous state. Processing both images through the matching network, the estimated response map $\hat{y} = f_{{\bf{w}}^{adapt}}(x,z)$ is obtained. 
The new position of the target can be found by finding the maximum position in the response map $\hat{y} \otimes h$, where $\otimes$ is an element-wise multiplication operator and $h$ is a cosine window function for penalizing large displacements. Scale variation of the target can be covered by using multiple size crops of $z$ matched with $x$. Scale changes are also penalized and damped by a constant to ensure smooth changes of target size over time. 

During the course of tracking, we keep a memory of the context images as ${\bf{z}}_{mem} = \{z_1,...,z_K\}$ along with the corresponding estimated response maps used for tracking $\hat{{\bf{y}}}_{mem} = \{\hat{y}_1, ..., \hat{y}_K \}$. We store a context image $z$ to the memory only if it is considered to be confident, where the maximum response value in the corresponding map $\hat{y}$ is over a certain threshold $\tau$. To update the appearance model of the target, we choose $M$ samples from this memory under the minimum entropy criterion on $\hat{{\bf{y}}}_{mem}$ as in \cite{meem} without replacement. This criterion is used to avoid ambiguous response maps where false positive samples may exist in the corresponding context image. Finding the response map with the minimum entropy can be defined as,
\begin{equation}\label{eq:entropy}
\argmin_{\hat{y}_i \in \hat{{\bf{y}}}_{mem}} -\sum_{\mathrm{p} \in \mathcal{P}}{ \rho (\hat{y}_i[\mathrm{p}]) \log (\rho (\hat{y}_i[\mathrm{p}]))},
\end{equation}
where $\mathrm{p}$ corresponds to a position in a set of all possible positions $\mathcal{P}$ in the response map and $\rho(\cdot)$ is the normalization function. Using the chosen $M$ appearance samples ${\bf{z}}_\delta$ , target-adaptive weights $w^{target}$ are obtained using the meta-learner network as in \eqref{eq:grad} and \eqref{eq:meta}, and then the matching network is updated as in \eqref{eq:update}, and it is used to track the object in subsequent frames. Since updating the model too frequently is unnecessary and cumbersome for the performance, we only update the model every $T$ frames as in other algorithms \cite{ECO}. The overall tracking process is described in Algorithm \ref{alg:track}.

\begin{figure}[t]
	\begin{center}
		\includegraphics[width=1.0\linewidth]{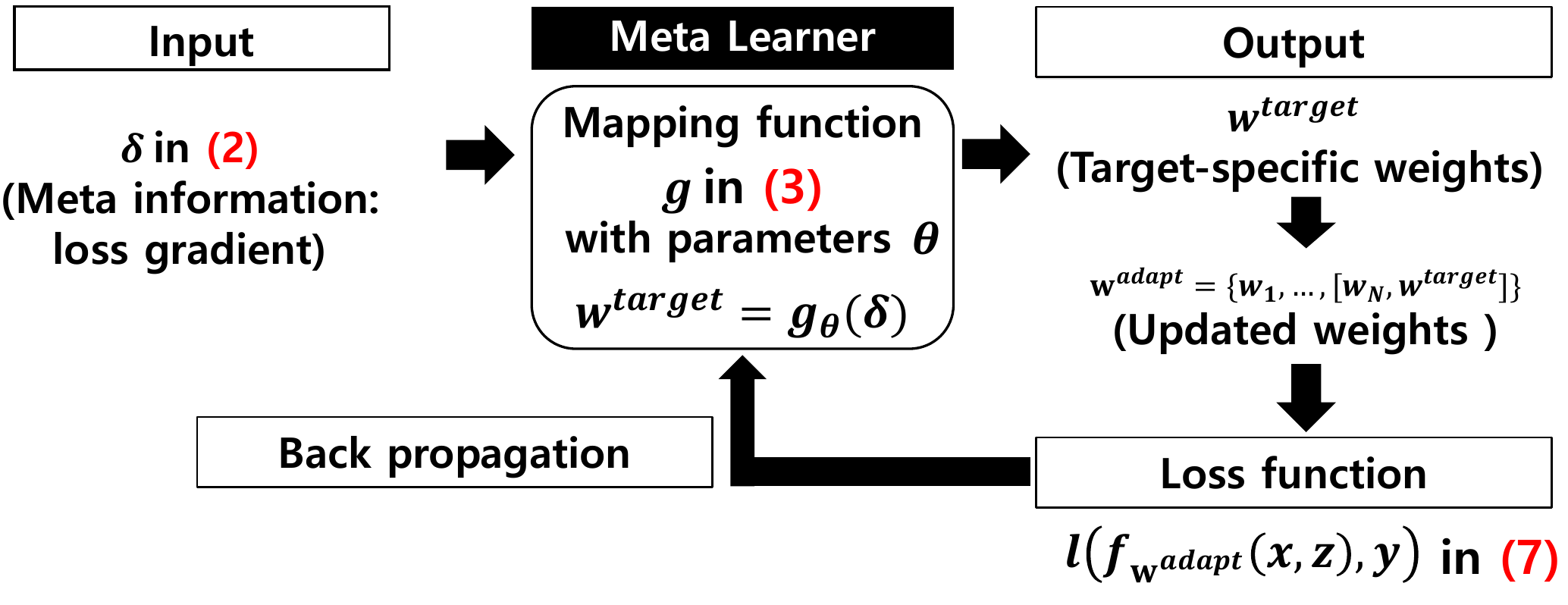}
	\end{center}
	\vspace{-3mm}
	\caption{{\textbf{Training scheme of meta-learner network.}} The meta-learner network uses loss gradients $\delta$ in \eqref{eq:grad} as meta information, derived from the matching network, which explains its own status in the current feature space \cite{metanet}. Then, the function $g(\cdot)$  in \eqref{eq:meta} learns the mapping from this loss gradient to adaptive weights $w^{target}$, which describe the target-specific feature space. The meta-learner network can be trained by minimizing the loss function in \eqref{eq:metaloss}, which measures how accurate the adaptive weights $w^{target}$ were at fitting new examples $\{z_1,...,z_{M'} \}$ correctly.}
	\label{fig:mln_training}
\vspace{-3mm}
\end{figure}

%-------------------------------------------------------------------------
\subsection{Network Implementation and Training} \label{sub:details}

%-------------------------------------------------------------------------
\subsubsection{Matching Network}

The matching network consists of a shared feature extractor CNN $\phi(\cdot)$, channel-wise attention step, feature $\ell^2$-normalization step, and cross-correlation step. For feature extraction, we use a CNN with $5$ convolutional layers and $2$ pooling layers of kernel size $3$ and stride $2$ are applied after the first two convolutional layers. Batch normalization layer is inserted after each convolutional layer. The overall structure of the CNN is analogous to \cite{siamfc}, where kernel size and input/output dimensions for each layer are {\small$w_1:11\times11\times3\times128$, $w_2:5\times5\times128\times256$, $w_3:3\times3\times256\times384$, $w_4:3\times3\times384\times256$, $w_5:1\times1\times256\times192$}. For inputs, we use a RGB image of size $127\times127\times3$ for $x$ and a RGB image of size $255\times255\times3$ for $z$, and the matching network produces a response map of size $17\times17$.

To train the matching network, we used ILSVRC 2015 \cite{imagenet} object detection from video dataset with additional training data from ILSVRC 2017 dataset, which contains objects of 30 classes in 4000 videos in the training set and 1314 videos in the validation set, with a total of 11566 independent object trajectories. Each frame of the video is annotated with bounding box notations of objects appearing in the video. We only used videos in the training set to train the matching network. At training time, pairs of $(x,z)$ are randomly sampled from an object trajectory in a chosen video clip. Then, a ground-truth response map $y \in {\{-1,+1\}}^{17 \times 17}$ is generated where the value is $+1$ at the position of the target and $-1$ otherwise.  For the loss function $\ell(f_{\bf{w}}(x,z),y)$, we use the logistic loss function defined as,
\begin{equation}\label{eq:loss}
\begin{split}
&\ell(f_{\bf{w}}(x,z),y) = {} \\
&\frac{1}{|\mathcal{P}|} \sum_{\mathrm{p}\in\mathcal{P}} \zeta(y[\mathrm{p}]) \cdot \log\left(1 + \exp(-f_{\bf{w}}(x,z)[\mathrm{p}] \cdot y[\mathrm{p}])\right),
\end{split}
\end{equation}
where $\mathrm{p}$ represents a position in the set of every possible positions $\mathcal{P}$ in the response map and $\zeta(y[\mathrm{p}])$ is a weighting function for alleviating label imbalance. The loss function is optimized with Adam  \cite{adam} optimizer with the learning rate of $10^{-4}$ using batch size of 8, and run for $95000$ iterations. 

%-------------------------------------------------------------------------
\subsubsection{Meta-Learner Network}

\begin{table*}[t]
	\resizebox{\textwidth}{!}{
		\begin{tabular}{@{}llllllllllllll@{}}
			\toprule[1.5pt]
			& \textbf{MLT}  & \textbf{SiamFC} & \textbf{StructSiam} & \textbf{DSiam} & \textbf{CFNet} & \textbf{SINT} & \textbf{SRDCF}  & \textbf{PTAV}  & \textbf{ECO-HC} & \textbf{STAPLE$_{\mathrm{CA}}$} & \textbf{BACF}  & \textbf{DSST} & \textbf{HDT}   \\ \midrule
			\textbf{OTB-2015}         & 0.611 & 0.582  & 0.621      & -     & 0.586 & 0.580 & 0.598 & 0.635 & \textbf{0.643} & 0.598 & 0.630 & 0.520     & 0.564 \\
			\textbf{OTB-2013}         & 0.621 & 0.607  & 0.638      & 0.642 & 0.611 & 0.635 & 0.626 & 0.663 & 0.652   & 0.621      & \textbf{0.678} & 0.554 & 0.603 \\
			\textbf{LaSOT} \smaller{Protocol I}  & \textbf{0.368} & 0.358  & 0.356      & 0.353 & 0.296 & 0.339 & 0.339 & 0.269 & 0.311   & 0.262      & 0.277 & 0.233 & -     \\
			\textbf{LaSOT} \smaller{Protocol II} & \textbf{0.345} & 0.336  & 0.335      & 0.333 & 0.275 & 0.314 & 0.314 & 0.250 & 0.304   & 0.238      & 0.259 & 0.207 & -     \\ \midrule
			\textbf{FPS}               & 48    & 58     & 45         & 45    & 43    & 4 & 5     & 25    & 60      & 35         & 35    & 24    & 10    \\ \bottomrule[1.5pt]
		\end{tabular}
	}
	\vspace{-2mm}
	\caption{ {\textbf{Quantitative results on OTB \cite{otb} and LaSOT \cite{LaSOT} datasets.}} MLT denotes the proposed algorithm. The proposed algorithm shows competitive performance on OTB datasets and outperforms other algorithms on large-scale LaSOT datasets, obtaining performance gains with the benefit of additional feature space provided by the meta-learner. AUC for OPE is used for the performance measures.}
	\label{table:comp}
	\vspace{-5mm}
\end{table*}

We then train the meta-learner network subsequent to pre-training the matching network. The meta-learner network $g_\theta(\cdot)$ consists of 3 fully-connected layers with 2 intermediate layers of $512$ units. Each intermediate layer is followed by a dropout layer with the keep probability of $0.7$ when training. For input, gradient $\delta$ of size $1\times1\times256\times192$ is used and output $w^{target}$ of size $1\times1\times256\times32$ is generated. These new kernels are used to update the weights of the matching network by concatenating $w^{target}$ to the kernels $w_5$ of the last layer of the Siamese matching network to provide the additional feature space needed for updates, resulting in new kernels $[w_5,w^{target}]$ of size $1\times1\times256\times(192+32)$. 

To train the meta-learner network, we use 1314 videos in the validation set of the ILSVRC video dataset. The training process is described hereafter. First, an anchor target image $x$ is randomly sampled from an object trajectory. Then, $M'$ context patches are randomly sampled from the same object's trajectory as in ${\bf{z}}_{reg}=\{z_1,...,z_{M'}\}$ where $M'\geq M$. Then $M$ patches are chosen from ${\bf{z}}_{reg}$ to form ${\bf{z}_\delta}$, where we can perform matching these samples with the target image $x$ to obtain gradient $\delta$ by \eqref{eq:grad} using generated binary response map $\tilde{y}_i$, assuming the target is located at the center of $z_i$. Standard data augmentation techniques (\emph{e.g.} horizontal flip, noise, Gaussian blur, translation) are applied when sampling $z_i$. We can train the meta-learner network $g_\theta(\delta)$ by minimizing the loss function with respect to parameter $\theta$:
\begin{equation}\label{eq:metaloss}
\begin{split}
\argmin_{\theta} \sum_{z_i \in {\bf{z}}_{reg}}\ell(f_{\textbf{w}^{adapt}}(x,z_i),y),~ \text{where}
\\ 
\textbf{w}^{adapt}=\{w_1,w_2,...,[w_N,g_\theta(\delta)]\}.
\end{split}
\end{equation}

\begin{table}[t]
	\centering
	\resizebox{0.45\textwidth}{!}{
		\begin{tabular}{@{}llll@{}}
			\toprule[1.5pt]
			& \textbf{\small{MLT}} & \makecell{\textbf{\small{MLT}\smaller{\textit{-mt}}}} & \makecell{\textbf{\small{MLT}-\scriptsize{\textit{mt+ft}}}} \\ 
			\midrule
			\textbf{\small{OTB-2015}} & \textbf{0.611} & 0.564 & 0.523 \\ 
			\textbf{\small{OTB-2013}} & \textbf{0.621} & 0.571 & 0.510 \\ 
			
			\textbf{\small{LaSOT}} \scriptsize{Protocol I} & \textbf{0.368} & 0.357 & 0.331 \\ 
			\textbf{\small{LaSOT}} \scriptsize{Protocol II} & \textbf{0.345} & 0.330 & 0.305 \\ 
			
			\textbf{\small{TC-128}} & \textbf{0.498} & 0.477 & 0.419 \\ 
			
			{\textbf{\small{UAV20L}} } & \textbf{0.435} & 0.366 & 0.342 \\ 
			
			{\textbf{\small{VOT-2016}} \scriptsize{Baseline}} & \textbf{0.537} & 0.514 & 0.517 \\ 
			{\textbf{\small{VOT-2016}} \scriptsize{Unsupervised}} & \textbf{0.421} & 0.412 & 0.411 \\ 
			\bottomrule[1.5pt]
		\end{tabular} 
	}
	\vspace{-1mm}
	\caption{\textbf{Internal comparison of tracking performance on OTB, LaSOT, TC-128, UAV20L and VOT-2016 datasets.} Proposed \textbf{MLT} shows consistent performance gains compared to \textbf{MLT\small{\textit{-mt}}} and \textbf{MLT\small{\textit{-mt+ft}}} throughout all datasets. For performance measures, AUC is shown for all experiments, with the exception of baseline experiment of VOT-2016 where A-R overlap score is shown.
		The best results were written in boldface.}
	\label{table:add}
	\vspace{-4mm}
\end{table}

Training the meta-learner network to generate new weights $w^{target}=g_\theta(\delta)$ that only fit examples in ${\bf{z}_\delta}$ (\emph{i.e.} $M'=M$ ) can lead the meta-learner network to generate weights that will make the matching network overfit to samples in ${\bf{z}_\delta}$. To prevent this overfitting problem, a regularization scheme is needed when training. For natural regularization, $M'=2M$ is chosen so that the weights can fit a larger set of examples ${\bf{z}}_{reg}$ instead of a smaller set ${\bf{z}_\delta}$. This encourages much better generalization properties for the matching network when tracking. For the experiments, $M=8$ and $M'=16$ are used and Adam optimizer with the learning rate of $10^{-4}$ with batches of 8 videos are used. Training is performed for $11000$ iterations.
Fig.\ref{fig:mln_training} shows the training scheme of the meta-learner network.

%-------------------------------------------------------------------------
\section{Experimental Results}

\begin{figure*}[t]
	\begin{center}
		\includegraphics[width=0.99\linewidth]{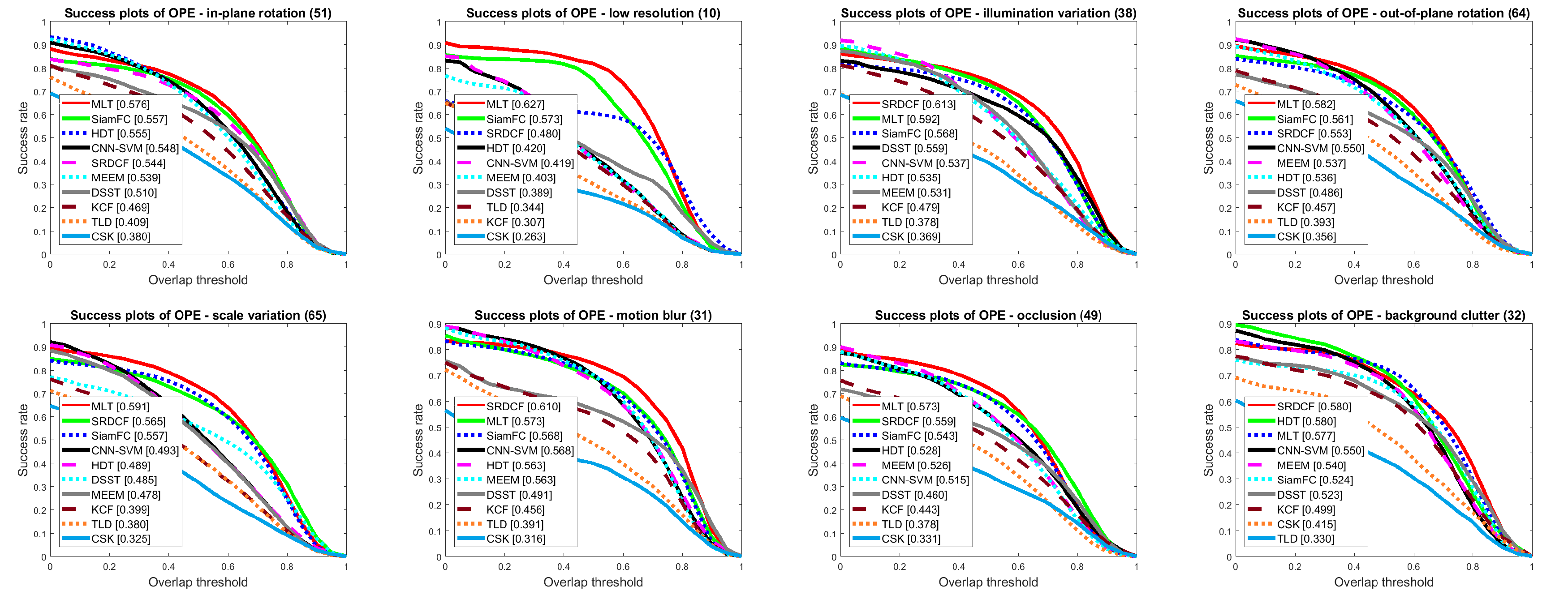}
	\end{center}
	\vspace{-2mm}
	\caption{{\textbf{Success plots for $8$ challenge attributes}} of the OTB-2015 dataset}
	\label{fig:attributes}
	\vspace{-4mm}
\end{figure*}

%-------------------------------------------------------------------------
\subsection{Evaluation Environment}
Our algorithm was implemented in Python using TensorFlow 1.8.0 \cite{tensorflow} library and executed on a system with Intel Core i7-4790K 4GHz CPU with 32GB of RAM with GeForce GTX TITAN X (Maxwell) GPU with 12GB of VRAM. The algorithm ran at an average of $48.1$ fps on $100$ videos in the OTB-2015 \cite{otb} dataset. We considered 3 scale variations of $[1.00, 1/1.035, 1.035]$ to adapt to the scale change of the target, where changes in scale are penalized by a constant of $0.97$ and damped by a constant of $0.59$. Cosine window $h$ was applied with the penalization factor of $0.25$. The meta-learner network updated the weights every $T=30$ frames, and threshold of $\tau=0.5$ was used for choosing confident samples. All parameters were fixed during the entire evaluation process for all datasets.

%-------------------------------------------------------------------------
\subsection{Experiments and Analysis}
\textbf{Object Tracking Benchmark (OTB)} \cite{otb} is a visual tracking benchmark that is widely used to evaluate the performance of a visual tracking algorithm. The dataset contains a total of 100 sequences and each is annotated frame-by-frame with bounding boxes and $11$ challenge attributes. OTB-2013 dataset contains $51$ sequences and the OTB-2015 dataset contains all $100$ sequences of the OTB dataset. As evaluation metric, we used OPE success rate evaluation metric that compares the predicted bounding boxes with the ground truth bounding boxes to obtain intersection over union (IoU) scores, and measure the proportion of predictions having larger score than a given varying threshold score value. The final score is calculated by measuring the area-under-curve (AUC) for each tracker. \textbf{Large-scale Single Object Tracking (LaSOT)} \cite{LaSOT} dataset is a recently introduced large-scale visual tracking dataset containing 1400 sequences with average length of 2512 frames (83 secs) and minimum of 1000 frames per sequence, with the total of 3.52 million frames where every frame is annotated with a bounding box annotation. It contains 70 object categories with each containing 20 sequences. Compared to OTB, LaSOT contains 14 times more sequences and 59 times the total number of frames, with more various object categories.  For evaluation protocols, \textit{Protocol I} employs all 1400 sequences for evaluation and \textit{Protocol II} uses the testing subset of 280 videos, where AUC of success plot is used for the performance metric for both protocols. 

We also perform internal comparisons on TC-128 \cite{tc128}, UAV20L \cite{uav123} and VOT-2016 \cite{vot} datasets to show the effectiveness of our meta-learner update scheme. VOT-2016 is the dataset used for the VOT Challenge \cite{vot} and it contains a total of 60 videos with bounding box annotations. Baseline experiment of the VOT dataset performs re-initialization of the tracker when it misses the target, while unsupervised experiment simply lets the tracker run from the first frame to the end. Temple Color-128 (TC-128) dataset \cite{tc128} contains 128 real-world color videos annotated with bounding boxes, with 11 challenge factors. UAV20L \cite{uav123} dataset contains 20 video sequences with an average length of 2933.5 frames, where some sequences have targets leaving the video frame (out-of-view). No explicit failure detection or re-detection scheme was used for all experiments.

%-------------------------------------------------------------------------
\subsubsection{Quantitative Analysis}

\noindent \textbf{Effect of meta-learner network: } 
We performed an internal comparison between the proposed tracker (\textbf{MLT}) and the baseline trackers \textbf{MLT\textit{-mt}} and \textbf{MLT\textit{-mt+ft}}, where \textbf{MLT\textit{-mt}} is a variant that has only the matching network with fixed weights without the meta-learner network, and \textbf{MLT\textit{-mt+ft}} performs on-line finetuning on \texttt{conv5} (kernel $w_5$) with training examples obtained while tracking. For fair comparison, the baseline trackers are pretrained on the whole ImageNet video detection dataset including the validation set, with the last convolutional layer of kernel size $1\times1\times256\times224$. For the \textbf{MLT\textit{-mt+ft}} method, we finetune the matching network every $50$ frames for $30$ iterations using the Adam optimizer with the learning rate of $10^{-3}$. As shown in Table \ref{table:add}, the meta-learner network improves the performance of the baseline matching network and produces better tracking results. \textbf{MLT} is consistently superior to \textbf{MLT\textit{-mt}} and \textbf{MLT\textit{-mt+ft}} in OTB, LaSOT, TC-128, UAV20L and VOT2016 datasets. The results demonstrate that the adaptive weights generated by meta-learner network are effective for inducing the customized feature space for each target and for resulting in accurate visual tracking, showing performance gains in all 5 datasets. Also, results of \textbf{MLT\textit{-mt+ft}} show that online finetuning without handpicked hyper-parameters and regularization scheme easily results in overfitting to handful of training samples, resulting in lower performance.

\begin{figure*}[t]
	\begin{center}
		\includegraphics[width=0.99\linewidth]{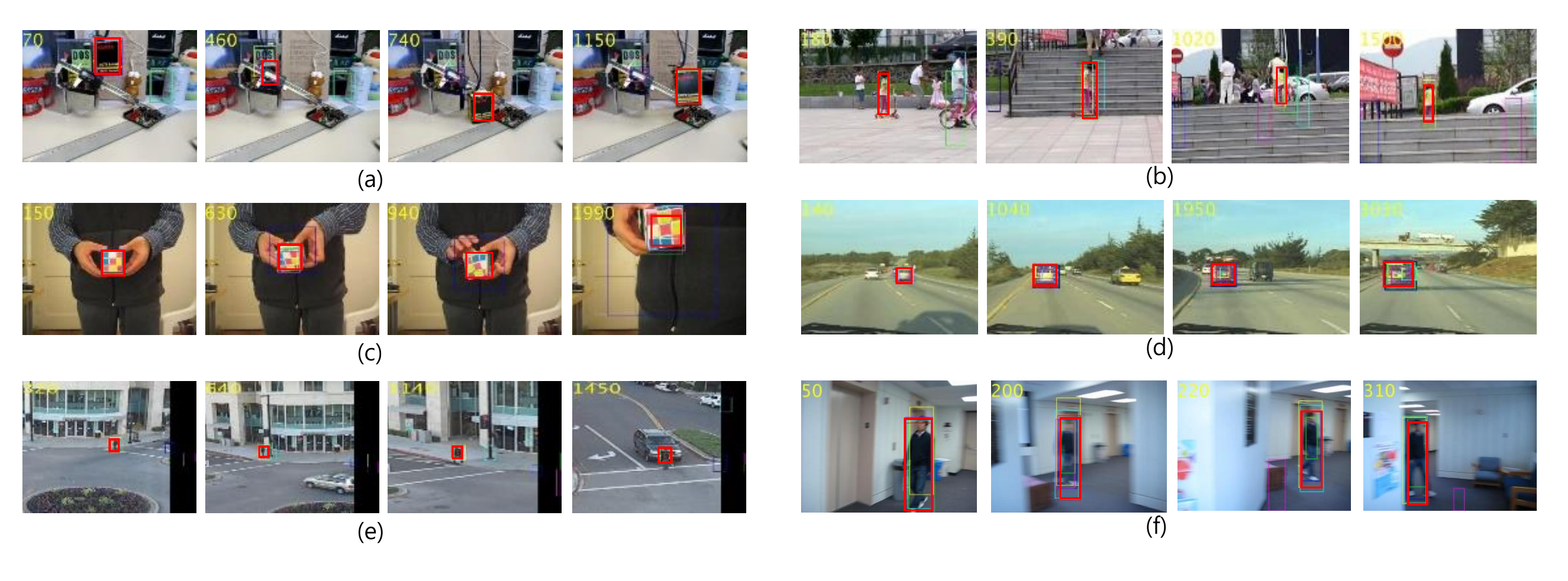}
	\end{center}
	\vspace{-5mm}
	\caption{{\textbf{Qualitative results.}} Tracking results for (a) \textit{box}, (b) \textit{girl2}, (c) \textit{rubik}, (d) \textit{car24}, (e) \textit{human3} and (f) \textit{blurBody} sequences. Green, Blue, Cyan, Yellow, Violet, and Red bounding boxes denote tracking results of SiamFC, SRDCF, HDT, CNN-SVM, DSST, and MLT, respectively.
		Yellow numbers on the top-left corners indicate frame numbers.
	}
	\label{fig:qualitative}
	\vspace{-4mm}
\end{figure*}

\begin{figure}[t]
	\centering
	\includegraphics[width=0.99\linewidth]{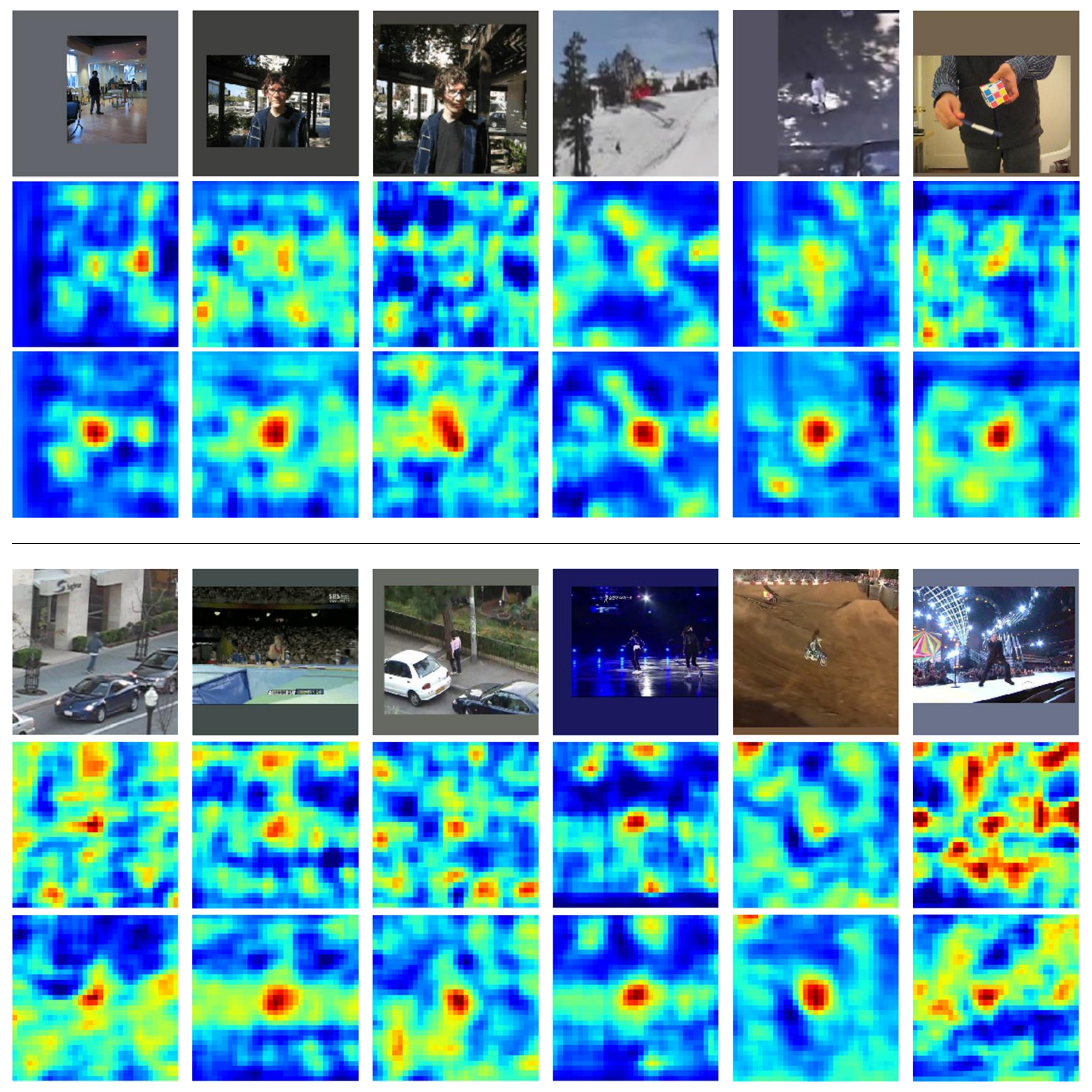}
	\caption{\textbf{Visualization for the effect of the target-specific feature space.} This shows some example image patches $z$ (1\textsuperscript{st} and 4\textsuperscript{th} row) with the changes in response maps $\hat{y}$ \textit{\textbf{before}} (2\textsuperscript{nd} and 5\textsuperscript{th} row) and \textit{\textbf{after}} (3\textsuperscript{rd} and 6\textsuperscript{th} row) applying our adaptive weights $w^{target}$ generated by our meta-learner.}
	\label{fig:visual}
	\vspace{-5mm}
\end{figure}

\noindent \textbf{Comparison with other trackers: } 
We compare our tracking algorithm MLT with 12 tracking algorithms on the OTB and LaSOT datasets, namely, SiamFC~\cite{siamfc}, StructSiam~\cite{stsiam}, DSiam~\cite{dsiam}, CFNet~\cite{cfnet}, SINT~\cite{siamis}, SRDCF~\cite{srdcf}, PTAV~\cite{ptav}, ECO-HC~\cite{ECO}, STAPLE\textsubscript{CA}~\cite{cacf}, BACF~\cite{bacf}, DSST~\cite{dsst} and HDT~\cite{hdt}, where most are real-time algorithms.
As shown in Table \ref{table:comp}, MLT achieves a competitive accuracy on OTB datasets compared to other tracking algorithms based on deep representation, and outperforms other algorithms on large-scale LaSOT experiments due to its robust meta-learner update. Especially, we were able to obtain a noticeable performance gain compared to SiamFC and its variants (StructSiam, DSiam and CFNet) on LaSOT where no variant was able to outperform the original SiamFC.

We also separately analyze the performance of MLT with respect to $8$ different attributes of OTB videos. 
Each video has a different attribute such as in-plane rotation, out-of-plane rotation, motion blur, low resolution, scale variation, illumination variation, background clutter and occlusion.
Fig. \ref{fig:attributes} shows that MLT is robust to low resolution, occlusion and scale variation and is competitive to the other trackers for most of the attributes.
In blurred low resolution images, the appearance of a target is frequently indistinguishable from that of other objects in the background. MLT can distinguish between these appearances by customizing features spaces for each target with the meta-learner network. Also, MLT can learn from negative examples from the background and handle occlusions better than other trackers.

%-------------------------------------------------------------------------
\subsubsection{Qualitative Analysis}
Fig.\ref{fig:qualitative} shows qualitative tracking results produced by SiamFC, SRDCF, HDT, CNN-SVM, DSST, and the proposed algorithm MLT.
All trackers were tested on all videos in the OTB-2015 dataset where tracking results for selected videos are shown in Fig.\ref{fig:qualitative} due to the limit in the length of the paper.
MLT robustly and accurately tracks the target in spite of several challenging conditions such as occlusion in {\textit{Box}} seq., pose variation in {\textit{Rubik}} seq., background clutter in {\textit{Human3}} seq., fast motion in {\textit{Girl2}} seq., and scale variation in {\textit{Car24}} seq..  
These qualitative tracking results demonstrate that the proposed MLT successfully exploited the power of the meta-learner network and utilized the adaptive weights customized for each target to improve tracking accuracy, without losing the generalization capabilities.   
For reference, we attach a supplementary video containing more qualitative results on the OTB-2015 dataset.

In addition, Fig.\ref{fig:visual} shows some examples of how the target-specific feature space modifies the response maps, thus demonstrating how the meta-learner can be beneficial in the task of visual tracking. We show example image patches $z$ where the target object fixed at the center of the image, with response maps before and after the target-specific feature space modification. The response maps show that the target-specific weights help the tracker to adapt to various target appearance changes and to locate the target, and are also effective in avoiding false positives by suppressing incorrect responses from distractors in the background.

%-------------------------------------------------------------------------
\section{Conclusion}
In this paper, we proposed a novel visual tracking algorithm based on the target-specific feature space constructed by the deep meta-learner network.
The proposed tracking algorithm adapts to the target appearance by generating the target-specific adaptive weights with the meta-learner network, where the matching network provides the meta-information gradients as a learning signal. Our algorithm aims to customize the feature space to discriminate a specific target appearance from the background in order to accurately track the target without overfitting.
Experimental results demonstrate that our algorithm achieves a noteworthy performance gain in visual tracking by using the proposed meta-learner network, achieving consistent performance gains on 5 tracking datasets including the large-scale tracking dataset LaSOT. 
Quantitatively and qualitatively the algorithm shows a competitive tracking performance on multiple visual tracking datasets with several challenging tracking conditions, compared to other visual tracking algorithms, while running at the real-time speed of $48$ fps.

\subsubsection*{Acknowledgments}
This work was supported by IITP grant funded by the Ministry of Science and ICT of Korea (No.2017-0-01780, The technology development for event recognition/relational reasoning and learning knowledge based system for video understanding).

{\small
	\bibliographystyle{ieeefullname}
	%\bibliography{egbib_mlt}

\begin{thebibliography}{10}\itemsep=-1pt
		
		\bibitem{tensorflow}
		Mart{\i}n Abadi, Ashish Agarwal, Paul Barham, Eugene Brevdo, Zhifeng Chen,
		Craig Citro, Greg~S Corrado, Andy Davis, Jeffrey Dean, Matthieu Devin, et~al.
		\newblock Tensorflow: Large-scale machine learning on heterogeneous distributed
		systems.
		\newblock {\em arXiv preprint arXiv:1603.04467}, 2016.
		
		\bibitem{siamfc}
		Luca Bertinetto, Jack Valmadre, Jo{\~a}o~F Henriques, Andrea Vedaldi, and
		Philip~HS Torr.
		\newblock Fully-convolutional siamese networks for object tracking.
		\newblock {\em arXiv preprint arXiv:1606.09549}, 2016.
		
		\bibitem{mosse}
		David~S Bolme, J~Ross Beveridge, Bruce~A Draper, and Yui~Man Lui.
		\newblock Visual object tracking using adaptive correlation filters.
		\newblock In {\em CVPR}, 2010.
		
		\bibitem{vot}
		Luka {\v{C}}ehovin, Ale{\v{s}} Leonardis, and Matej Kristan.
		\newblock Visual object tracking performance measures revisited.
		\newblock {\em IEEE TIP}, 25(3):1261--1274, 2016.
		
		\bibitem{ECO}
		Martin Danelljan, Goutam Bhat, Fahad Khan, and Michael Felsberg.
		\newblock Eco: Efficient convolution operators for tracking.
		\newblock In {\em CVPR}, 2017.
		
		\bibitem{dsst}
		Martin Danelljan, Gustav H{\"a}ger, Fahad Khan, and Michael Felsberg.
		\newblock Accurate scale estimation for robust visual tracking.
		\newblock In {\em BMVC}, 2014.
		
		\bibitem{deepsrdcf}
		Martin Danelljan, Gustav Hager, Fahad Shahbaz~Khan, and Michael Felsberg.
		\newblock Convolutional features for correlation filter based visual tracking.
		\newblock In {\em ICCV Workshop}, 2015.
		
		\bibitem{srdcf}
		Martin Danelljan, Gustav Hager, Fahad Shahbaz~Khan, and Michael Felsberg.
		\newblock Learning spatially regularized correlation filters for visual
		tracking.
		\newblock In {\em CVPR}, 2015.
		
		\bibitem{ccot}
		Martin Danelljan, Andreas Robinson, Fahad~Shahbaz Khan, and Michael Felsberg.
		\newblock Beyond correlation filters: Learning continuous convolution operators
		for visual tracking.
		\newblock In {\em ECCV}, 2016.
		
		\bibitem{color}
		Martin Danelljan, Fahad Shahbaz~Khan, Michael Felsberg, and Joost Van~de
		Weijer.
		\newblock Adaptive color attributes for real-time visual tracking.
		\newblock In {\em CVPR}, 2014.
		
		\bibitem{siamtriplet}
		Xingping Dong and Jianbing Shen.
		\newblock Triplet loss in siamese network for object tracking.
		\newblock In {\em ECCV}, 2018.
		
		\bibitem{LaSOT}
		Heng Fan, Liting Lin, Fan Yang, Peng Chu, Ge Deng, Sijia Yu, Hexin Bai, Yong
		Xu, Chunyuan Liao, and Haibin Ling.
		\newblock Lasot: {A} high-quality benchmark for large-scale single object
		tracking.
		\newblock {\em arXiv preprint arXiv:/1809.07845}, 2018.
		
		\bibitem{ptav}
		Heng Fan and Haibin Ling.
		\newblock Parallel tracking and verifying: A framework for real-time and high
		accuracy visual tracking.
		\newblock In {\em ICCV}, 2017.
		
		\bibitem{maml}
		Chelsea Finn, Pieter Abbeel, and Sergey Levine.
		\newblock Model-agnostic meta-learning for fast adaptation of deep networks.
		\newblock In {\em ICML}, 2017.
		
		\bibitem{dsiam}
		Qing Guo, Wei Feng, Ce Zhou, Rui Huang, Liang Wan, and Song Wang.
		\newblock Learning dynamic siamese network for visual object tracking.
		\newblock In {\em ICCV}, 2017.
		
		\bibitem{matchnet}
		Xufeng Han, Thomas Leung, Yangqing Jia, Rahul Sukthankar, and Alexander~C Berg.
		\newblock Matchnet: Unifying feature and metric learning for patch-based
		matching.
		\newblock In {\em CVPR}, 2015.
		
		\bibitem{Struck}
		Sam Hare, Amir Saffari, and Philip Torr.
		\newblock Struck: Structured output tracking with kernels.
		\newblock In {\em ICCV}, 2011.
		
		\bibitem{sasiam}
		Anfeng He, Chong Luo, Xinmei Tian, and Wenjun Zeng.
		\newblock A twofold siamese network for real-time object tracking.
		\newblock In {\em CVPR}, 2018.
		
		\bibitem{100fps}
		David Held, Sebastian Thrun, and Silvio Savarese.
		\newblock Learning to track at 100 fps with deep regression networks.
		\newblock In {\em ECCV}, 2016.
		
		\bibitem{KCF}
		Jo{\~a}o~F Henriques, Rui Caseiro, Pedro Martins, and Jorge Batista.
		\newblock High-speed tracking with kernelized correlation filters.
		\newblock {\em IEEE TPAMI}, 37(3):583--596, 2015.
		
		\bibitem{cnnsvm}
		Seunghoon Hong, Tackgeun You, Suha Kwak, and Bohyung Han.
		\newblock Online tracking by learning discriminative saliency map with
		convolutional neural network.
		\newblock In {\em ICML}, 2015.
		
		\bibitem{Muster}
		Zhibin Hong, Zhe Chen, Chaohui Wang, Xue Mei, Danil Prokhorov, and Dacheng Tao.
		\newblock Multi-store tracker (muster): A cognitive psychology inspired
		approach to object tracking.
		\newblock In {\em CVPR}, 2015.
		
		\bibitem{TLD}
		Zdenek Kalal, Krystian Mikolajczyk, and Jiri Matas.
		\newblock Tracking-learning-detection.
		\newblock {\em IEEE TPAMI}, 34(7):1409--1422, 2012.
		
		\bibitem{bacf}
		Hamed Kiani~Galoogahi, Ashton Fagg, and Simon Lucey.
		\newblock Learning background-aware correlation filters for visual tracking.
		\newblock In {\em ICCV}, 2017.
		
		\bibitem{adam}
		Diederik Kingma and Jimmy Ba.
		\newblock Adam: A method for stochastic optimization.
		\newblock {\em ICLR}, 2015.
		
		\bibitem{alexnet}
		Alex Krizhevsky, Ilya Sutskever, and Geoffrey\texttt{}~E Hinton.
		\newblock Image{N}et classification with deep convolutional neural networks.
		\newblock In {\em NIPS}, 2012.
		
		\bibitem{lenet}
		Yann LeCun, L{\'e}on Bottou, Yoshua Bengio, and Patrick Haffner.
		\newblock Gradient-based learning applied to document recognition.
		\newblock {\em Proceedings of the IEEE}, 86(11):2278--2324, 1998.
		
		\bibitem{siamrpn}
		Bo Li, Junjie Yan, Wei Wu, Zheng Zhu, and Xiaolin Hu.
		\newblock High performance visual tracking with siamese region proposal
		network.
		\newblock In {\em CVPR}, 2018.
		
		\bibitem{tc128}
		Pengpeng Liang, Erik Blasch, and Haibin Ling.
		\newblock Encoding color information for visual tracking: Algorithms and
		benchmark.
		\newblock {\em IEEE TIP}, 24(12):5630--5644, 2015.
		
		\bibitem{csrdcf}
		Alan Lukezic, Tomas Vojir, Luka Cehovin~Zajc, Jiri Matas, and Matej Kristan.
		\newblock Discriminative correlation filter with channel and spatial
		reliability.
		\newblock In {\em CVPR}, 2017.
		
		\bibitem{convcorr}
		Chao Ma, Jia-Bin Huang, Xiaokang Yang, and Ming-Hsuan Yang.
		\newblock Hierarchical convolutional features for visual tracking.
		\newblock In {\em CVPR}, 2015.
		
		\bibitem{L1}
		Xue Mei and Haibin Ling.
		\newblock Robust visual tracking using l1 minimization.
		\newblock In {\em ICCV}, 2009.
		
		\bibitem{uav123}
		Matthias Mueller, Neil Smith, and Bernard Ghanem.
		\newblock A benchmark and simulator for uav tracking.
		\newblock In {\em ECCV}, 2016.
		
		\bibitem{cacf}
		Matthias Mueller, Neil Smith, and Bernard Ghanem.
		\newblock Context-aware correlation filter tracking.
		\newblock In {\em CVPR}, 2017.
		
		\bibitem{metanet}
		Tsendsuren Munkhdalai and Hong Yu.
		\newblock Meta networks.
		\newblock In {\em ICML}, 2017.
		
		\bibitem{mdnet}
		Hyeonseob Nam and Bohyung Han.
		\newblock Learning multi-domain convolutional neural networks for visual
		tracking.
		\newblock In {\em CVPR}, 2015.
		
		\bibitem{metatracker}
		Eunbyung Park and Alexander~C Berg.
		\newblock Meta-tracker: Fast and robust online adaptation for visual object
		trackers.
		\newblock In {\em ECCV}, 2018.
		
		\bibitem{hdt}
		Yuankai Qi, Shengping Zhang, Lei Qin, Hongxun Yao, Qingming Huang, Jongwoo Lim,
		and Ming-Hsuan Yang.
		\newblock Hedged deep tracking.
		\newblock In {\em CVPR}, 2016.
		
		\bibitem{metalstm}
		Sachin Ravi and Hugo Larochelle.
		\newblock Optimization as a model for few-shot learning.
		\newblock In {\em ICML}, 2016.
		
		\bibitem{frcnn}
		Shaoqing Ren, Kaiming He, Ross Girshick, and Jian Sun.
		\newblock Faster r-cnn: Towards real-time object detection with region proposal
		networks.
		\newblock In {\em NIPS}, 2015.
		
		\bibitem{IVT}
		David Ross, Jongwoo Lim, Ruei-Sung Lin, and Ming-Hsuan Yang.
		\newblock Incremental learning for robust visual tracking.
		\newblock {\em IJCV}, 77(1–3):125--141, 2008.
		
		\bibitem{imagenet}
		Olga Russakovsky, Jia Deng, Hao Su, Jonathan Krause, Sanjeev Satheesh, Sean Ma,
		Zhiheng Huang, Andrej Karpathy, Aditya Khosla, Michael Bernstein, et~al.
		\newblock Imagenet large scale visual recognition challenge.
		\newblock {\em IJCV}, 115(3):211--252, 2015.
		
		\bibitem{vgg}
		Karen Simonyan and Andrew Zisserman.
		\newblock Very deep convolutional networks for large-scale image recognition.
		\newblock {\em arXiv preprint, arXiv:1409.1556}, 2014.
		
		\bibitem{crest}
		Yibing Song, Chao Ma, Lijun Gong, Jiawei Zhang, Rynson Lau, and Ming-Hsuan
		Yang.
		\newblock Crest: Convolutional residual learning for visual tracking.
		\newblock In {\em ICCV}, 2017.
		
		\bibitem{siamis}
		Ran Tao, Efstratios Gavves, and Arnold W~M Smeulders.
		\newblock Siamese instance search for tracking.
		\newblock In {\em CVPR}, 2016.
		
		\bibitem{cfnet}
		Jack Valmadre, Luca Bertinetto, Joao Henriques, Andrea Vedaldi, and Philip
		H.~S. Torr.
		\newblock End-to-end representation learning for correlation filter based
		tracking.
		\newblock In {\em CVPR}, 2017.
		
		\bibitem{matchos}
		Oriol Vinyals, Charles Blundell, Tim Lillicrap, Koray Kavukcuoglu, and Daan
		Wierstra.
		\newblock Matching networks for one shot learning.
		\newblock In {\em NIPS}, 2016.
		
		\bibitem{fctrack}
		Lijun Wang, Wanli Ouyang, Xiaogang Wang, and Huchuan Lu.
		\newblock Visual tracking with fully convolutional networks.
		\newblock In {\em ICCV}, 2015.
		
		\bibitem{stct}
		Lijun Wang, Wanli Ouyang, Xiaogang Wang, and Huchuan Lu.
		\newblock Stct: Sequentially training convolutional networks for visual
		tracking.
		\newblock In {\em CVPR}, 2016.
		
		\bibitem{nipstrack}
		Naiyan Wang and Dit-Yan Yeung.
		\newblock Learning a deep compact image representation for visual tracking.
		\newblock In {\em NIPS}, 2013.
		
		\bibitem{otb}
		Yi Wu, Jongwoo Lim, and Ming-Hsuan Yang.
		\newblock Object tracking benchmark.
		\newblock {\em IEEE TPAMI}, 37(9):1834--1848, 2015.
		
		\bibitem{mccnn}
		Jure {\v{Z}}bontar and Yann LeCun.
		\newblock Stereo matching by training a convolutional neural network to compare
		image patches.
		\newblock {\em JMLR}, 17(1):2287--2318, 2016.
		
		\bibitem{meem}
		Jianming Zhang, Shugao Ma, and Stan Sclaroff.
		\newblock {MEEM:} robust tracking via multiple experts using entropy
		minimization.
		\newblock In {\em ECCV}, 2014.
		
		\bibitem{CST}
		Tianzhu Zhang, Adel Bibi, and Bernard Ghanem.
		\newblock In defense of sparse tracking: Circulant sparse tracker.
		\newblock In {\em CVPR}, 2016.
		
		\bibitem{MTT}
		T. Zhang, B. Ghanem, S. Liu, and N. Ahuja.
		\newblock Robust visual tracking via multi-task sparse learning.
		\newblock In {\em CVPR}, 2012.
		
		\bibitem{MCPF}
		Tianzhu Zhang, Changsheng Xu, and Ming-Hsuan Yang.
		\newblock Multi-task correlation particle filter for robust object tracking.
		\newblock In {\em CVPR}, 2017.
		
		\bibitem{stsiam}
		Yunhua Zhang, Lijun Wang, Jinqing Qi, Dong Wang, Mengyang Feng, and Huchuan Lu.
		\newblock Structured siamese network for real-time visual tracking.
		\newblock In {\em ECCV}, 2018.
		
		\bibitem{dasiamrpn}
		Zheng Zhu, Qiang Wang, Bo Li, Wei Wu, Junjie Yan, and Weiming Hu.
		\newblock Distractor-aware siamese networks for visual object tracking.
		\newblock In {\em ECCV}, 2018.
		
	\end{thebibliography}
	
}

\end{document}